\documentstyle[
here,twoside,cite,fleqn]{article}
\title{%
Rigid Body
Structure and Motion 
From Two-Frame Point-Correspondences
Under Perspective Projection. 
}
\author{Mieczys{\l}aw A. K{\l}opotek\\
{\footnotesize\sl Institute of Computer Science, 
Polish Academy of Sciences}\\
{\footnotesize\sl PL 01-248 Warsaw, 5 Jana Kazimierza St.,}\\
{\footnotesize\sl   e-mail:
klopotek{@}ipipan.waw.pl} }

\newcommand{\V}{\vec}
\date{}

\newcommand{\Bem}[1]{}

\newcommand{\Zitat}[1]{~\cite{#1}}

\begin{document}

\maketitle

{
\footnotesize
{\bf Abstract.}\ \
This paper is concerned with possibility of 
recovery of motion and structure parameters
from multiframes under perspective projection when only points on a rigid
body are traced. Free (unrestricted and uncontrolled) pattern of motion
between frames is assumed. 
The major question is how many points and/or how many frames are necessary
for the task. It has been shown in an earlier paper \Zitat{Klopotek:95b}
that for orthogonal projection two frames are insufficient for the task.
The paper demonstrates that, under perspective projection,
that total uncertainty about relative position
of focal point versus projection plane 
makes the recovery of structure and motion 
from two frames impossible. 
}

\section{Introduction}

Recovery of a three-dimensional structure  from a single view  of even the
simplest scene consisting
of a single object has been viewed as heavily underconstrained 
\Zitat{Lee:88}. On the other hand, availability of multiframes 
may provide with 
 additional constraints which may lead to solvability of the problem 
\Zitat{Lee:88}. Therefore, the problem of recovery of rigid bodies
from multiframes has been studied in the past.
E.g. in the domain of orthogonal projections,
Lee\Zitat{Lee:88} has
been concerned with rigid bodies 
consisting of two traceable points 
rotating around a fixed
direction, Klopotek\Zitat{Klopotek:92g} has studied  rigid bodies 
consisting of three traceable points 
subject to free (unrestricted) motion, and Klopotek\Zitat{Klopotek:92}
investigated  rigid bodies 
consisting of two traceable points connected by a smooth 3D curve 
subject to free (unrestricted) motion.  In the domain  of perspective
projections, Weng\Zitat{Weng:92} deals with the recovery of motion 
and structure of rigid bodies consisting only of straight lines (13 of them).
  Roach and Aggarwal~\Zitat{Roach:79}~\Zitat{Roach:80} 
researched on motion and structure recovery tracing points under perspective 
projection assuming static 
scene and moving camera. They showed that five points in two views are needed 
to recover the structure and motion parameters. Their solution involved a 
system of 18 highly non-linear equations. Nagel~\Zitat{Nagel:81} proposed a 
simplified equation system by separating solution for the translation vector 
and the rotation matrix, with rotation matrix being determined by a system of 
3 equations in three motion parameters.
Wang et al.\Zitat{Wang:91} studied bodies consisting of four points and a
line. 
Azarbayejami and Pentland\Zitat{Azarbayejami:95} reviewed and studied 
problems of structure and motion recovery under 
unknown (but fixed) focal length.

Generally, much effort has been devoted  to reducing the number
of frames and traceable features (points, lines). This is understandable as on
the one hand this reduces the effort required for tracing features and on the
other hand there are more features left for validation and/or improvement of
error resistance. 

It is generally known that  the amount of
information provided by the frames shall at least balance the number of
degrees of freedom involved. 
If the balance of degrees of freedom and of information is not achieved then
results concerning structure and motion may be totally ambiguous. 
This may prove to be extremely difficult to notice when performing numerical
computations, as especially for perspective projection most methods of
recovery of structure and motion involve complex non-linear multivariate
equation systems which may yield a unique solution (due to numerical
round-offs or imprecision of observations) even if such a solution does not
exist. This was demonstrated e.g. in \Zitat{Klopotek:94c} for the
four-points-and-a-line 
algorithm from \Zitat{Wang:91} 
  for two frames for perspective projection
 (no. of degrees of freedom exceeding the information available) \\

On the other hand it appears that information available from frames
may be divided into two
categories: new information and redundant information \Zitat{Klopotek:95b}.
If the balance of degrees of freedom and of information is achieved 
but the  balance of degrees of freedom and of new information is not achieved 
then
results concerning structure and motion may be partially ambiguous, as
demonstrated 
 in \Zitat{Klopotek:95b} for four-points problem for two frames for
orthogonal projection (no increase in new information over three points due
to a shift towards redundant information). 

This paper demonstrates that perspective projections are also prune to the
risk of emergence of
redundant information. In section 2 we recall the
problem of emergence of information redundancy as observed for orthogonal
projection. Then in section 3 the situation for perspective projection is
described where information redundancy emerges. In section 4 we discuss
briefly how to prevent this information split and how to make use of it. 

The paper ends with a brief discussion and some concluding remarks.

\section{Split of Information Under Orthogonal Projection}

\subsection{Degrees of Freedom for Orthogonal Projection}
    
Under orthogonal projection, 
  each point of the body introduces 3 df in the 
first frame   minus one df for the whole body as there exists 
no possibility of determining the initial depth of the
 body in the space. The 
motion introduces for each subsequent 
frame 5 df only (three for rotations and two for translation), because the
motion in the direction orthogonal to the
projection plane has no impact on the image. In general, with p points forming
the rigid body traced over k frames we have 
$-1+3*p+5*(k-1)$ degrees of freedom.
On the other hand, within each image each traced point provides us with two 
pieces of information: its x and its y position within the frame. Hence 
we have at most 
$ k*2*p $ pieces of information available from k images.
    Thus we need at least to have the balance 
\begin{equation}
           -1+3*p+5*(k-1)      \le   k*2*p
\end{equation}
to achieve recoverability. 
\Bem{
    Let us consider some combinations of parameters:
\begin{itemize}
\item for k=3 frames, p=3 points we get
$-1+3*p+5*(k-1)=18 = k*2*p=18$ 
\item for k=2 frames, p=4 points we get
$-1+3*p+5*(k-1)=-1+12+5=16 =  k*2*p=2*2*4=16$ 
\end{itemize}
Please notice that with p=2 points we have no chance of recovery of structure
and motion whatever the number of frames, because always:
\begin{equation}
           -1+3*p+5*(k-1)=-1+6+5*(k-1)=5*k > k*2*2
\end{equation}
}
\subsection{Information Redundancy}
As we can derive from the above equations, if the number of traced points is
equal 4, then the amount of information may be sufficient to recover structure
and motion from two frames. However,  let us assume that we managed to match 
 two frames with a 3D object consisting of 4 or more points, that is 
 we construct an object in space and find positions of two projection planes
such that the projection of the object on two planes gives the two observed
frames. Then, as shown in in\Zitat{Klopotek:95b}, we can rotate one of the
frames along a specially selected axis by any angle just to obtain still
another different 3D object that also matches the two frames.

 This implies that  forth and subsequent points do carry only one
piece of information in the second image instead of two. Hence there exists no
possibility of complete recovery of 3-D structure from two images. 

The question seems at this point to be justified what happens with the one
piece of information left unused. As shown in
\Zitat{Klopotek:95b}, they
can be exploited for solving the problem of correct assignment of identities
of points in two consecutive frames. 

\section{Information Redundancy Preventing Structure and Motion Recovery
from Two Frames
 under Perspective Projection}

In most papers concerning recovery of structure and motion from multiframes
the detailed knowledge of the geometry of the optical system of the camera is
assumed, that is the precise position of the projection center point with
respect to the projection plane is known. However, this does not need to be
always the case. If we take the image e.g. from a household video camera (a
fine one  with  auto-focusing)  then  the  relative  position  of 
projection center
point and the image plane is not only unknown but also varying over time. If
images from a photographic  camera of unknown type are available only then
the information on relative position of image plane and focal point is also
inaccessible. Below we demonstrate that under these circumstances it is
impossible to recover structure and motion from two  frames whatever number of
traceable points is taken. First we demonstrate that it is not the  overall
number of degrees of freedom is the obstacle. Then we show that informational
redundancy occurs consuming the information necessary for recovery of
structure and motion.

\subsection{Degrees of Freedom}

Let us now consider the degrees of freedom for the perspective projection
if we assume that the relative position (in space) of the focal 
point with respect to the projection plane is not known and may vary over
time.

  Each point of the body introduces 3 df in the 
first frame   minus one df for the whole body as there exists 
no possibility of determining the scaling of the whole body.
Additionally we have 3df due to the uncertainty of the
location of the focal point.
 The 
motion introduces for each subsequent 
frame 9 df (three for rotations and three for translation of the projection
plane plus three for translation of the focal point). In general, with p
points forming the rigid body traced over k frames we have  then 
$-1+3*p+3+9*(k-1)$ degrees of freedom.
On the other hand, within each image each traced point provides us with two 
pieces of information: its x and its y position within the frame. Hence 
we have at most 
$ k*2*p$ pieces of information available from k images.
    Thus we need at least to have the balance 
\begin{equation}
           -1+3*p+3+9*(k-1)      \le   k*2*p
\end{equation}
to achieve recoverability. 
    Let us consider some combinations of parameters:
\begin{itemize}
\item for k=2 frames, p= 10 points we get
$-1+3*p+3+9*(k-1)=41 > k*2*p=40$ 
\item for k=2 frames, p= 11 points we get
$-1+3*p+3+9*(k-1)=44 = k*2*p=44$ 
\item for k=2 frames, p=7 points we get
$-1+3*p+3+9*(k-1)=32 >  k*2*p=28      $ 
\item for k=3 frames, p=7 points we get
$-1+3*p+3+9*(k-1)=41 <  k*2*p=42      $ 
\item for k=3 frames, p=6 points we get
$-1+3*p+3+9*(k-1)=38 >  k*2*p=36      $ 
\item for k=4 frames, p=6 points we get
$-1+3*p+3+9*(k-1)=47 <  k*2*p=48      $ 
\item for k=8 frames, p=5 points we get
$-1+3*p+3+9*(k-1)=80 =  k*2*p=80      $ 
\end{itemize}

The above (in)equalities tell us that to recover structure and motion from 5
traceable points, we would need 8 images (frames), with 7 traceable points
we need 3 frames, and to recover from two frames we would need 11 points - if
we take the balance of degrees of freedom and the amount of information.

If we have only p=4 traceable points, then we get the number of degrees of
freedom equal to -1+3*4+9*(k-1)=9k+2, whereas the amount of information
is equal to k*2*4=8k, which is always less then the number of degrees of
freedom. This means that if we trace only four points, we can never recover
structure and motion whatever number of frames is available.

\subsection{Emerging Information Redundancy}

We will demonstrate in this paper, however, 
 that it is impossible to recover structure and
motion from two frames only because the information stemming from points
beyond first seven is redundant.\\

\begin{figure}
\setlength\unitlength{1cm}
 \begin{picture}(10,5)(-1,-7)
  \input FRAME2.PIC
 \end{picture}


\caption{Frame 2. Positions of points $F_1''$, $B''$, $D''$, $F''$ and
$H''$ are assumed to be unknown.}
\label{AbbDrei}
\end{figure}
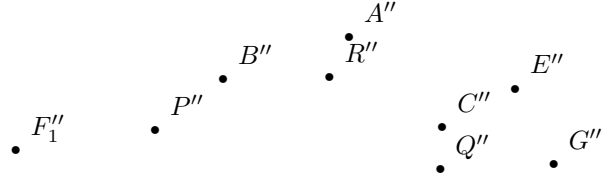

 Let us consider a rigid body consisting of seven traceable points 
   P, Q, R, 
          A, C, E and G. We shall assume that no four of them are coplanar.
(On treatment of four coplanar points compare \Zitat{Klopotek:92,%
Klopotek:90f}). 
Let us assume that their two projections are available, frame 1 with    P', 
Q',
 
 R', A', C', E' and G' (see Fig.\ref{AbbZwei}),  and frame 2  with  P", Q", 
R",
 A", C", E" and G"  (see Fig.\ref{AbbDrei}).
 What we claim now is that if we have another
point Z with its projection Z' in the first frame, then we can draw a line
$z"$ in the second frame on which the projection of Z onto the second frame
must
lie. In other words given Z', the point Z" has only one intrinsic degree of
freedom to be located in the second frame. This means also that knowledge of
the location of Z" contributes only one piece of information to the recovery
of structure and motion. As each new point introduces 3df and provides 2
pieces of information in the first and only one in the second frame (=3df in
all), then addition of any further point does not contribute anything to the
solution of structure and motion problem.

\begin{figure}[H]
\setlength\unitlength{1cm}
 \begin{picture}(10,5)(-5,0)
  \input FRAME1.PIC 
 \end{picture}


\caption{Frame 1}
\label{AbbZwei}
\end{figure}

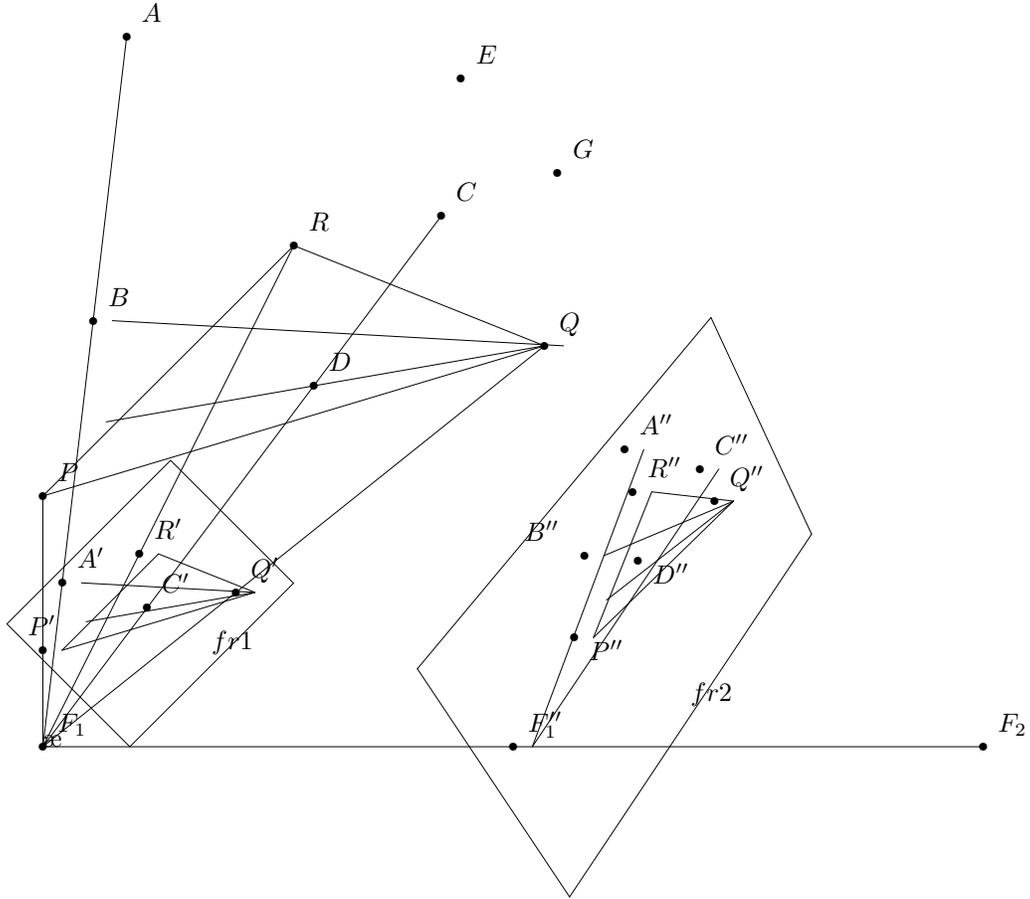
\begin{figure}
\setlength\unitlength{1cm}
 \begin{picture}(10,10)
  \input SCENE1.PIC
  \input SCENE2.PIC
 \end{picture}

  \vspace{2cm}

\caption{A scene with seven traced points over two frames}
\label{AbbEins}
\end{figure}

To demonstrate the validity of our claim let us imagine that not the traced
body moves but rather the projection plane and the focal point. Let $F_1$ be
the intrinsic 
 position of the focal point of the first frame and $F_2$ that of the focal
point
of the second frame (see Fig.\ref{AbbEins}). Under this convention we define
$F_1"$ as the
projection of point $F_1$ onto the second frame. Let us define straight lines
$a=F_1A=F_1A'$, $c=F_1C=F_1C'$, $e=F_1E=F_1E'$, $g=F_1G=F_1G'$. Let us
consider the plane PQR. Let $B$, $D$, $F$, and $G$ be
the
points of intersection of lines $a,c,e,g$ and the plane PQR respectively. Let 
\\
 $B"$, $D"$, $F"$, $H"$ be projections of 
 $B$, $D$, $F$, and $G$ onto the second frame
(with respect to its focal point $F_2$). Let Z be
the eighth point of the rigid body and we define the line $z=F_1Z$ and the
points $Z'$ - projection of $Z$ on frame 1, 
,$Z"$ - projection of $Z$ on frame 2, 
$Z_{PQR}$ - intersection of $PQR$ with $z$,
$Z"_{PQR}$ - projection of $Z_{PQR}$ on frame 2, 
in analogous way.

 To show the validity of our
claim we demonstrate first, that given in the first frame: $P',Q',R',A',C',
E', G',$ and in the second frame: $P", Q", R", A", C",D",E"$ we can identify
$F_1"$ in the second frame. 

Then we show that given additionally $Z'$ in the
first frame, we can identify $Z"_{PQR}$ in the second frame. But then we have
clearly identified the line $z"=F_1"Z"_{PQR}$ which will complete the 
proof.

{

}

\subsubsection{Basic Geometrical Facts}

\begin{figure}[H]
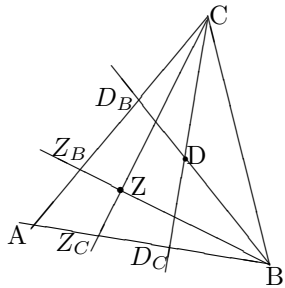

\begin{center}
\input ABCD_CO.PIC
\end{center}
\caption{Illustration of double quotient coordinates of a point in plane}
\label{figdq}
\end{figure}

So let us first recall the well-known theorem on double quotient (DQ) which
says the following (see Fig.\ref{figdq}: if points A,B,C,D are collinear and
A', B', C', D' are
their perspective projections onto a plane (perspective projection preserves
collinearity), then the following holds:
\begin{equation} \label{DQ}
 DQ(A,B,C,D)=
\frac{AC}{AD}:\frac{BC}{BD}=\frac{A'C'}{A'D'}:\frac{B'C'}{B'D'}
= DQ(A',B',C',D')
\end{equation}

\begin{figure}[H]
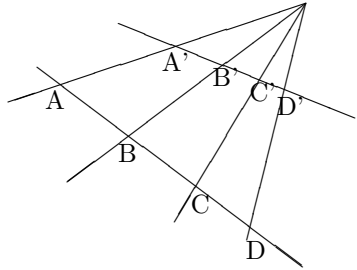

\begin{center}
\input DQ.PIC
\end{center}
\caption{Illustration of double quotient property}
\label{figabcdco}
\end{figure}

This actually means the following for the operation of perspective projection
onto a frame fr with a focal point F: 
Given three collinear points A,B,C
and their (perspective) projections A',B',C' onto a plane, and given a forth
point
D on the line AB, then we can identify the position of projection D' of D 
on the line A'D', even if we  know neither the position of F nor that of the
frame fr in space with respect to points A,B,C. 

What is more,  given  four  coplanar   points  A,B,C,D  (no  three 
collinear) together
with their
projections  A',B',C',D', and given a point Z in the plane ABC, then we can
uniquely determine the position Z' of the projection of Z onto the plane 
A'B'C',  even if we  know neither the position of the focal point F nor that
of the frame fr in space with respect to points A,B,C,D. 
This is straight forward to achieve via equation (\ref{DQ}).  
Let $D_B$ denote the intersection of lines DB and BC, 
let $D_C$ denote the intersection of lines DC and AB, 
Let $Z_B$ denote the intersection of lines ZB and BC, 
let $Z_C$ denote the intersection of lines ZC and AB
(see Fig.\ref{figabcdco}. 
Also,    
let $D_B'$ denote the intersection of lines D'B' and B'C', 
let $D_C'$ denote the intersection of lines D'C' and A'B', 
It's obvious that $D_B'$ is the projection of $D_B$, and 
 $D_C'$ is the projection of $D_C$.

Obviously, $A,C,D_B,Z_B$ are  collinear,  and   $A,B,D_C,Z_C$  are 
collinear,
and $A',B',D_C'$ are collinear and $A',C',D_B'$ are also collinear.
Hence $Z_B',Z_C'$, the projections of $Z_B,Z_C$ resp. are easily located.
Now $Z'$ is easily located as the intersection of lines $B'Z_B'$ and
$C'Z_C'$. 

Note that double quotients 
$\frac{AD_C}{BD_C}:\frac{AZ_C}{BZ_C}$ and
$\frac{AD_B}{CD_B}:\frac{AZ_B}{CZ_D}$
$\frac{AC}{AD}:\frac{BC}{BD}$ may be considered as "coordinates" of Z in the
ABCD coordination system, preserved under any sequence of perspective
projections. 

Furthermore, if three points A',B',C' (projections of some points A,B,C) are
not collinear, then there exists always a series of projections with respect
to suitably chosen focal points and projection planes such that 
in the last frame, with $A ^n$,  $B ^n$,  $C ^n$ being images of $A',B',C'$,
  $A ^nB ^n$ is orthogonal to  $A ^nC ^n$ and line segments 
  $A ^nB ^n$ and               $A ^nC ^n$ are of unit length.

Last not least, if points A,zB,C,D are coplanar, then lines AB and CD are
either parallel or they intersect.

These well known facts from elementary geometry prove very fruitful
when applied to our task. Let us turn back to the situation 
depicted in Fi.\ref{AbbEins}.

\subsubsection{Locating Projected Focal Point $F_1"$}

Let us now identify the position of the projection $F_1"$ onto the frame 2
of the focal point $F_1$ of the frame 1. We know only relative positions of
points
$R',P',Q',A',C',E', G'$ relatively in the frame 1, and
$R'',P'',Q'',A'',C'',E'', G''$ relatively in the frame 2. We assume that
we have already transformed by a sequence of perspective projections frame 2
in such a way that line segments $R'',P''$,$R''Q''$ are orthogonal and both of
unit lengths
$R"$ be the origin of coordinate system, $R"Q"$ the X-axis, $R"P"$ the Y-axis.
 As the sequence of projections from original
frame2 to a transformed one is known a double quotient preserving, we lose no
information and can always locate  $F_1"$ in the original frame 2.

Given the information, it is easy to locate points:
$A_P=$ intersection of $A'P'$ and $R'Q'$, $A_Q=$ intersection of $A'Q'$ and
$R'P'$, 
$C_P=$ intersection of $C'P'$ and $R'Q'$, $C_Q=$ intersection of $C'Q'$ and
$R'P'$, 
$E_P=$ intersection of $E'P'$ and $R'Q'$, $E_Q=$ intersection of $E'Q'$ and
$R'P'$,
 $G_P=$ intersection of $G'P'$ and $R'Q'$, $G_Q=$ intersection of $G'Q'$ and
$R'P'$. \\

We can also calculate the double quotients:
\begin{eqnarray}
qCQ=\frac{||RA_Q||}{||PA_Q||}:\frac{||RC_Q||}{||PC_Q||},&
qCP=\frac{||RA_P||}{||QA_P||}:\frac{||RC_P||}{||QC_P||}, \nonumber\\
qEQ=\frac{||RA_Q||}{||PA_Q||}:\frac{||RE_Q||}{||PE_Q||},&
qEP=\frac{||RA_P||}{||QA_P||}:\frac{||RE_P||}{||QE_P||}, \nonumber\\
qGQ=\frac{||RA_Q||}{||PA_Q||}:\frac{||RG_Q||}{||PG_Q||},&
qGP=\frac{||RA_P||}{||QA_P||}:\frac{||RG_P||}{||QG_P||} \label{edq} 
\end{eqnarray}

What are the constraints on the positions of  $R'',P'',Q'',A'',C'',E'', G''$ 
in the frame 2 ? First of all, all the  lines $A"B"$, $C"D"$, $E"F"$, $G"H"$
must intersect at a single point that is at $F_1"$
(because  by definition  $AB$, $CD$, $EF$, $GH$
 intersect at a single point that is at $F_1$). This leads us to the following
equation system:%
\begin{eqnarray}
t\cdot \V{A"B"}=t_1\cdot \V{C"D"} \label{e1}\\
t\cdot \V{A"B"}=t_2\cdot \V{E"F"} \label{e2}\\
t\cdot \V{A"B"}=t_3\cdot \V{G"H"} \label{e3}
\end{eqnarray}

We can solve  the three  linear equation systems 
(\ref{e1}),  (\ref{e2}),  (\ref{e3}) for $t$, 
and from comparison of $t$ from the first two equation systems we get:

\begin{eqnarray}
0=   (C".x-E".x)*(B".x-A".x)*(D".y-C".y)*(F".y-E".y)\nonumber\\
+    (C".y-E".y)*(B".y-A".y)*(D".x-C".x)*(F".x-E".x)\nonumber\\
-    (C".x-A".x)*(B".y-A".y)*(D".y-C".y)*(F".x-E".x)\nonumber\\
-    (C".y-A".y)*(B".x-A".x)*(D".x-C".x)*(F".y-E".y)\nonumber\\
+    (E".x-A".x)*(B".y-A".y)*(D".x-C".x)*(F".y-E".y)\nonumber\\
+    (E".y-A".y)*(B".x-A".x)*(D".y-C".y)*(F".x-E".x) \label{x2} 
\end{eqnarray}

\begin{figure}[H]
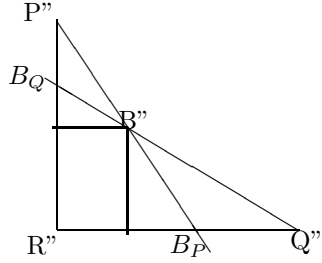

\begin{center}
\input TALES.PIC
\end{center}
\caption{Illustration to equation 
                                  (9)}
\label{figtales}
\end{figure}

Let us introduce "coordinate" points of $B",D",F",H"$ as follows:
$B_P=$ intersection of $B"P"$ and $R"Q"$, $B_Q=$ intersection of $B"Q"$ and
$R"P"$, 
$D_P=$ intersection of $D"P"$ and $R"Q"$, $D_Q=$ intersection of $D"Q"$ and
$R"P"$, 
$F_P=$ intersection of $F"P"$ and $R"Q"$, $F_Q=$ intersection of $F"Q"$ and
$R"P"$, 
$H_P=$ intersection of $H"P"$ and $R"Q"$, $H_Q=$ intersection of $H"Q"$ and
$R"P"$,

We see easily that (see Fig.\ref{figtales})

\begin{eqnarray}
B".x=(1/B_Q.y-1)/((1/B_P.x)*(1/B_Q.y)-1) \nonumber \\
B".y=(1/B_P.x-1)/((1/B_P.x)*(1/B_Q.y)-1) \label{x3a}
\end{eqnarray}

and so forth for other auxiliary points $D"$, $F"$ and $H"$.

Observe that,  due to our assumption of R",Q",P"
establishing the coordinate system, we have also (see
equation (\ref{edq})):

\begin{eqnarray}
1/D_P.x=(1-qCP)+qCP/B_P.x \nonumber \\
1/D_Q.y=(1-qCQ)+qCQ/B_Q.y \nonumber \\
1/F_P.x=(1-qEP)+qEP/B_P.x \nonumber \\ 
1/F_Q.y=(1-qEQ)+qEQ/B_Q.y \label{x3b}
\end{eqnarray}

Substitution of equations (\ref{x3a}) and then  (\ref{x3b}) into 
equation (\ref{x2}) results in a polynomial equation 
in only two unknowns: $\frac{1}{B_P.x}$ and   $\frac{1}{B_Q.y}$.

Transforming analogously (\ref{e1}) and  (\ref{e3}) by first 
eliminating  $t$, we finally get another, independent equation in 
the same two unknowns: $\frac{1}{B_P.x}$ and   $\frac{1}{B_Q.y}$.

Note that both are cubic in  $\frac{1}{B_Q.y}$ (and  also  $\frac{1}{B_P.x}$).
By multiplying both equations with factors standing in front of 
 $(\frac{1}{B_Q.y})^3$ in the other equation and then subtracting both we get
a quadratic equation in  $\frac{1}{B_Q.y}$, easily solved symbolically. 
It is easily observed, that one of the solutions would always be the
degenerate solution $\frac{1}{B_Q.y}=1$ (meaning a collapse of B",D",F",H"
and $F_1"$ onto P"), so we will always take the other one (just 
having a unique
solution at that moment).\\

Then we substitute this symbolic result into one of the two ("cubic") 
equations substituting for  $\frac{1}{B_Q.y}$ and getting a one variable
polynomial equation in  $\frac{1}{B_P.x}$, solvable by conventional methods.\\

In a simulated experimental setting we  had observations:

\begin{center}
\begin{tabular}{||r|l|r|r||}
\hline \hline
Frame & Point & .x & .y \\
\hline
1    &  R'    &  1.00   &  3.29   \\
1    &  Q'    &  3.00   & 11.50   \\
1    &  P'    &  1.84   &  5.53   \\
1    &  A'    &  1.82   &  6.05   \\
1    &  C'    &  1.63   &  5.42   \\
1    &  E'    &  2.09   &  7.51   \\
1    &  G'    &  1.74   &  5.56   \\
\hline
\hline
\end{tabular}
\begin{tabular}{||r|l|r|r||}
\hline \hline
Frame & Point & .x & .y \\
\hline
2    &  R"    &  0.00   &   0.00  \\
2    &  Q"    &  1.00   &   0.00  \\
2    &  P"    &  0.00   &   1.00  \\
2    &  A"    & 23.80   &  33.95  \\
2    &  C"    & 21.20   &  30.26  \\
2    &  E"    & 15.59   &  20.73  \\
2    &  G"    & 16.92   &  24.92  \\
\hline
\hline
\end{tabular}
\end{center}

We got  one of  solutions    $\frac{1}{B_P.x}$=1.43 and
$F_1"(-16,-23)$, 

The behavior of the final polynomial in  $\frac{1}{B_P.x}$ was as follows:

\begin{center}
\begin{tabular}{||r|r||}
\hline \hline
 $\frac{1}{B_P.x}$  & Polynomial \\
\hline
1.33     &  -0.85    \\
1.35     &  -0.68    \\
1.37     &  -0.52    \\
1.39     &  -0.34    \\
1.41     &  -0.18    \\
1.43     &   0.00    \\
1.45     &   0.19    \\
1.47     &   0.31    \\
1.49     &   0.60    \\
1.51     &   0.83    \\
1.53     &  1.07     \\
\hline
\hline
\end{tabular}
\end{center}

\subsubsection{Locating line z'}

If we knew now the position of the point Z' (projection of Z) in the first
frame, we could calculate proper double quotient in frame 1,
find the projection of intersection point of $F_1Z$ and $PQR$ onto frame 
2 and then draw the line connecting this point with $F_1"$, which is just the
line $z"$ we were looking for.  
 Q.E.D.

\subsubsection{Freedom of Shape of the Identified "Rigid" Multipoint Object}

The results of the previous subsection  mean that 
no matter how many  points are
given in two images, it is always possible to find countless fittings of
frames yielding different objects (not only by size, but also by
shape!) that may be source of of both projections. As we stated earlier, there
are four degrees of freedom unusable by any fitting procedure. We will show
below the effects of two of these degrees of freedom only, because imagination
of the other two is far more complicated.

\begin{figure}[H]
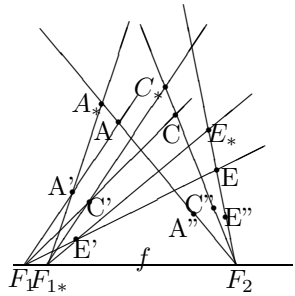

\begin{center}
\input F1MOVE.PIC
\end{center}
\caption{Illustration impact of motion of $F_1$ along $f$ 
on "recovered" positions of points A,B,C in space}
\label{figf1move}
\end{figure}

Assume that we fit together two frames with n  point correspondences each.
E.g. A' and A" are projections of a point A in space
(see Fig.\ref{figf1move}). Let $f$ be the line
joining focal points $F_1F_2$ (in 3D space). Obviously, points $A'$, $A"$,
$F_1$ and $F_2$ and line $f$ are coplanar. $A$ is the intersection of $F_1A'$
and $F_2A"$. Now let us "move" $F_1$ along $f$ to another location, say
$F_{1*}$ (The projection frame 1 is left as it was in space). Obviously, 
 points $A'$, $A"$,
$F_{1*}$ and $F_2$  are also coplanar, and 
 $F_1A'$ and $F_2A"$ (most probably) intersect at a point $A_*$.  The same
happens with the other (n-1) traced points: after moving $F_1$ the "rays"
starting at $F_{1*}$ and $F_2$ and passing through traced points in frame 1
and frame 2 resp. still intersect, but at different points. That is, another
3D object may also have given the same two projections. And this new object is
usually pretty different from the previous one.

Very same manipulations can be done "moving" $F_2$ along $f$, giving other
different 3D objects.\\

\section{Exploiting and Preventing  Information Redundancy}

As in case of orthogonal projection, we may put now the question 
 what happens with the one piece of information
 of the eighth point left unused. Let us consider what this
info means
geometrically. Given the first seven points, for each further
point, if we know its image in the first frame,  we can identify the line 
on which it lies in the second frame.  This means a point Z
with its image Z' in the first frame must have its image lying on a
concrete
line z" in the second frame. But if Z" does not lie on the pre-specified
line? Than two things may have happened. Either Z is not a part of a rigid
body containing P, Q , R, S, T, U, W , or \dots the identities of P", Q",
etc. have been assigned incorrectly. 

But the latter means that if we have a set of projected points S1 and a set of
projected points S2 of which we know that they are projections of a set of
points belonging to a rigid body, but the identities are not ascribed, then
we may be capable of assigning identity relations among points of the set S1
and the set S2. For this purpose we may select eight points from the set
S1 and try allocating to them points of the set S2. In all, if n is the
cardinality of the set S2 (equal to the cardinality of the set S1) we may have
to try
$  \frac{n!}{(n-8)!}$ 
combinations of points. (In case of n=8 we have 8! combinations).
First seven  points are then used to identify the line on which the
eighth point
should lie in the second frame, and the distance between the line and the real
position of projected point will be used to evaluate the goodness (or in fact
the badness) of fit. The identity assignment minimizing the distance may be
considered as the best. It is, however, easily seen that the task may be
prohibitive. It is advisable to use additional information (e.g. substructures
of visible connections between points) to bind the complexity. 

Under these circumstances the question seems to be justified to what extent
the geometry of the optical system must be known in order to enable recovery
of structure and motion from two frames. In \Zitat{Klopotek:94c} it has been
demonstrated that the knowledge of exact position the focal point relatively
to the  projection plane imposes the following requirement on the balance of
degrees of freedom and the amount of information:
\begin{equation}
                -1+3*p+6*(k-1) \le k*2*p
\end{equation}  

With 
p=5 points and k=2 frames we get $-1+3*p+6*(k-1)=20= k*2*p=20$. Papers
\Zitat{Roach:79,Roach:80} deal with recovery in that case.

Can we weaken the geometrical requirements ? 
First let us consider the case where the relative position of projection plane
and the focal point is unknown, but fixed. We have then that 
  each point of the body introduces 3 df in the 
first frame   minus one df for the whole body as there exists 
no possibility of determining the scaling of the whole body.
Additionally we have 3df due to the uncertainty of the
location of the focal point.
With p
points forming the rigid body traced over k frames we have  
$-1+3*p+3+6*(k-1)$ degrees of freedom. To achieve the balance we require:
$$-1+3*p+3+6*(k-1) \le k*2*p$$ 
If we fix k at level 2, then we require $   3*p+8   \le 4*p$ meaning $p\ge 8$.
Hence we get the same trouble with  the seven-point-limit. 

Now what if we know the  relative position of the focal point and of the
projection plane  up to the distance between them (that is that the focal
point may only move towards and away from the projection plane along a fixed
axis, a requirement fulfilled by typical modern cameras with unsupervised
autofocusing) ? Each point of the body introduces 3 df in the 
first frame   minus one df for the whole body as there exists 
no possibility of determining the scaling of the whole body.
Additionally we have 1df due to the uncertainty of the
location of the focal point.
With p
points forming the rigid body traced over k frames we have  
$-1+3*p+1+7*(k-1)$ degrees of freedom. To achieve the balance we require:
$$-1+3*p+1+7*(k-1) \le k*2*p$$ 
If we fix k at level 2, then we require $   3*p+7   \le 4*p$ meaning $p\ge 7$.
In this case we have just met the seven-point limit.

\section{Discussion}

In this paper we have demonstrated that for perspective projection
of rigid
 bodies in some situations 
 at least three frames are necessary to recover
structure and motion. 
From a degrees-of-freedom argument it became visible that the amount of
information that two frames with eleven traced points may provide enough
information to recover structure and motion from two frames. However, it has
been demonstrated that this is impossible because the rigid body assumption
imposes internal dependence between the point projections so that 
information
provided by the eighth point and any further traced point cannot be
consumed for purposes of recovery of structure and motion.

We need to stress that purely geometrical properties of "points" have been
considered. In practical settings we have generally to handle errors in
positioning points in the frame raster. If we now assume that
there exists a (possibly stochastic) dependence between
measurement errors
and the distance between (at least some) observed points and the camera, then
we may have a clue how to
recover the distance object-camera and may overcome the phenomenon
demonstrated in this paper. But if the error of measurement does not depend 
on the distance from the camera, but on other factors, then there exists no
possibility to recover the complete set of structure and motion parameters
from two frames under perspective projections (from purely geometrical
point-dependent clues).\\
  
Instead,  eight or more points over two frames may solve
identification problem of points between consecutive frames or alternatively
the problem of belonging to the same rigid body. That is, in the first case,
if we have two frames with 8    (or more) points each and we know that these
points belong to the same rigid body, but we do not know the exact point to
point correspondence, then we can exploit the unused information (not
consumable for recovery of structure and motion) for purposes of
identification of point-to-point correspondences. Alternatively, in the second
case, when we have sets of points in two frames where the
point-to-point-correspondence between frames is known, then we can exploit the
 unused information (not
consumable for recovery of structure and motion) to decide, which points
belong to the same rigid body.

It is worth mentioning at this point that several papers claimed possibility
of recovery of structure and motion from two frames (using less then 7 
points)
\Zitat{Roach:79,Roach:80,Weng:92,Wang:91}. It must be stressed that in those
papers the complete knowledge of geometry of the optical system is assumed.
In that case, clearly, the number of degrees of freedom is different
and statements about the necessary number of points and frames are different.
 In
\Zitat{Klopotek:94c} we have demonstrated that the structure and motion
recovery method proposed in \Zitat{Wang:91} (two frames, four points and a
line) is not correct due to unbalanced
degrees of freedom. On the other hand, under such conditions, recovery for
five points and two frames \Zitat{Roach:79,Roach:80} is possible.

Though recovery of structure and motion of rigid bodies consisting only of
(a limited) number of traced points may seem to be a simplistic task, it is
still of practical relevance. E.g. we live in times of rapidly growing image
databases especially in criminology. Review by hand of such databases may
prove prohibitive and hence some clues restricting the search space
significantly are of importance. Claims have been raised that some simple
 measurements of spatial structure of a few points on the surface of face may
be sufficient to identify the suspect. The important question is then: how
many images (e.g. from a video camera of a bank security system) are needed,
and how many features are to be traced to recover the 3D structure of points
of interest. This study demonstrates the impact of knowledge of geometrical
structure of the optical system. Strict knowledge allows to recover the
structure from 2 images and 5 traced  points \Zitat{Nagel:81}.
If we know the geometry up to the distance image plane
- focal point 
, then we can still
work with 2 images, however with 7 traceable points. And if we are totally
ignorant of the geometry 
, at least 3 images are required - with  7 traced points.

\section{Conclusions}

\begin{itemize}
\item It is impossible to recover structure or motion from two frames whatever
number of traced points is available, if there is complete uncertainty about
relative position of projection plane and focal point from frame to frame. For
recovery at least 3 images are needed. The same is true even if this relative
position does not  change over time but is unknown.
\item If a rigid  body consists of at least eight points, then 
we can solve the problem of point tracing for any two consecutive frames alone
from knowledge which points of two frames belong to the body (without explicit
knowledge of point-to-point correspondence)
\item Alternatively, if a rigid  body consists of at least eight points,
then we can solve the problem of belonging to a rigid body for any
two consecutive frames alone
from  explicit
knowledge of point-to-point correspondence.
\item If  we know the geometry of the optical system up to the distance image
plane - focal point (e.g. from a camera with autofocusing), then we can
recover structure and motion 
work with 2 images, however with 7 traceable points.
\item  Strict knowledge of the geometry of the optical system  allows to
recover
the structure from 2 images and 5 traced  points or 3 images and 4 points
\Zitat{Klopotek:94c}.
\end{itemize}

\newcommand{\LitStelle}[2]{\bibitem{#1} }   


\begin{thebibliography}{99}
\bibitem{Azarbayejami:95}
A. Azarbayejami, A.P. Pentland:
Recursive estimation of motion, structure and focal length,
{\it IEEE Trans. PAMI}, 17(1995)562-575.
\LitStelle{Klopotek:89d}{Klopotek, 1989d} 
K{\l}opotek M.A.: Physical space in  reconstruction  of 
    moving curves, [in:] 
Proc. 
National  CIR'89  (Cybernetics,  Intelligence, 
    Development) Conference, Siedlce (Poland) 18-20.9.1989,
 Vol. I, 55-71 (1989)
\LitStelle{Klopotek:90f}{Klopotek, 1990f}  
K{\l}opotek M.A.: 3-D-Shape  reconstruction  of  moving   curved 
    objects,
    [in:] V.  Miszalok  Ed.: 
    {\it MedTech'89 Medical Imaging},  
     Proc. SPIE 1357,  29-39 (1990)
\bibitem{Klopotek:92g} 
K{\l}opotek M.A.
A simple method of  recovering  3D-curves from  
    multiframes, 
{\it Archiwum     Informatyki     Teoretycznej     i 
     Stosowanej},  Vol.4, No. 1-4,  103-110 (1992).
%
\bibitem{Klopotek:94c} 
{K{\l}opotek M.A.}: 
A comment on "Analysis of video image sequences using point and line
correspondences", 
{\it Pattern Recognition} Vol. 28 No. 2, pp. 283-292, 1995
\LitStelle{Klopotek:95b}{Klopotek, 1995b}  
{K{\l}opotek M.A.}: 
Distribution of Degrees of Freedom over
Structure and Motion of Rigid Bodies, 
{\it Machine Graphics \& Vision}, Vol 4 No 1/2, pp. 83-100 (1995)
\LitStelle{Klopotek:92}{Klopotek, 1992} 
K{\l}opotek M.A. :  
     Reconstruction of 3-D rigid smooth curves moving free when    
     two traceable points only are available)
     {\it Machine  Graphics  and 
     Vision}, Vol. I, nos 1/2, 1992, 392-405
%
\bibitem{Lee:88} 
Lee C.H.:  Interpreting  image  curve  from  multiframes,   
   {\it Artificial Intelligence } 35(, 145-164 1988)
%
\bibitem{Nagel:81} 
Nagel H.H.: Representation of moving rigid objects based on visual 
observations, Computer 29-39 (August 1981).
%
\bibitem{Roach:79} 
Roach J.W., Aggarwal J.K.: Computer tracking of objects moving in space, 
{\it IEEE 
Trans. Pattern Anal. Mach.Intell } 1,127-135 (1979)
%
\bibitem{Roach:80} 
Roach J.W., Aggarwal J.K.: Determining the  movement of objects from a
sequence 
of images, 
{\it IEEE 
Trans. Pattern Anal. Mach.Intell. } 3,554-562(1980). 
%
%
\bibitem{Wang:91}
Wang Y.F., Karandikar N., Aggarwal J.K.: 
Analysis of video image sequences using point and line correspondences, 
{\it Pattern Recognition } Vol.24 No.11, 1065-1085 (1991)
%
\bibitem{Weng:92}
Weng J., Huang T.S., Ahuja N.: 
Motion and structure from line correspondences: Closed form solution,
{\it IEEE Transactions on Pattern Analysis and Machine Intelligence } 
Vol.14, No.3, 318-336(1992)
\end{thebibliography}
\end{document}